\documentclass[runningheads]{llncs}
\usepackage[T1]{fontenc}
\usepackage{graphicx}
\usepackage{graphicx}
\usepackage{amsmath}
\usepackage{amssymb}
\usepackage{booktabs}
\usepackage{multirow}
\usepackage{enumitem}
\usepackage{placeins}

\begin{document}
\title{Symmetry Defense Against \\ CNN Adversarial Perturbation Attacks}
%
%
\author{Blerta Lindqvist\orcidID{0000-0002-4950-2250}}
\institute{Aalto University, Espoo, Finland \\
\email{blerta.lindqvist@aalto.fi}}


%
\maketitle              
\begin{abstract}

This paper uses symmetry to make Convolutional Neural Network classifiers (CNNs) robust against adversarial perturbation attacks. Such attacks add perturbation to original images to generate adversarial images that fool classifiers such as road sign classifiers of autonomous vehicles.
Although symmetry is a pervasive aspect of the natural world, CNNs are unable to handle symmetry well. For example, a CNN can classify an image differently from its mirror image.
For an adversarial image that misclassifies with a wrong label $l_w$, CNN inability to handle symmetry means that a symmetric adversarial image can classify differently from the wrong label $l_w$. 
Further than that, we find that the classification of a symmetric adversarial image reverts to the correct label.
To classify an image when adversaries are unaware of the defense, we apply symmetry to the image and use the classification label of the symmetric image.
To classify an image when adversaries are aware of the defense, we use mirror symmetry and pixel inversion symmetry to form a symmetry group. We apply all the group symmetries to the image and decide on the output label based on the agreement of any two of the classification labels of the symmetry images.
Adaptive attacks fail because they need to rely on loss functions that use conflicting CNN output values for symmetric images.
Without attack knowledge, the proposed symmetry defense succeeds against both gradient-based and random-search attacks, with up to near-default accuracies for ImageNet. The defense even improves the classification accuracy of original images.

\end{abstract}

\section{Introduction}
\label{sec:introduction}

Despite achieving state-of-the-art status in computer vision~\cite{he2016deep,krizhevsky2012imagenet}, convolutional neural network classifiers (CNNs) lack adversarial robustness because they can classify imperceptibly perturbed images incorrectly~\cite{carlini2017towards,goodfellow6572explaining,madry2017towards,szegedy2013intriguing}. One of the first and still undefeated defenses against adversarial perturbation attacks is adversarial training~(AT)~\cite{kurakin2016adversarial,madry2017towards,szegedy2013intriguing}, which uses adversarial images in training. However, AT reliance on attack knowledge during training~\cite{madry2017towards} is a significant drawback since such knowledge might not be available.

Although engineered to incorporate symmetries such as horizontal flipping, translations, and rotations, CNNs lack invariance with respect to these symmetries~\cite{engstrom2019exploring} in the classification of datasets such as ImageNet~\cite{deng2009imagenet}, CIFAR10~\cite{krizhevsky2009cifar}, MNIST~\cite{lecun1998mnist}. The CNN lack of invariance means that CNNs can classify images differently after they have been horizontally flipped, or even slightly shifted or rotated~\cite{azulay2019deep,engstrom2019exploring}. Furthermore, CNNs only provide approximate translation invariance~\cite{azulay2019deep,bouchacourt2021grounding,engstrom2019exploring,kayhan2020translation} and are unable to learn invariances with respect to symmetries such as rotation and horizontal flipping with data augmentation~\cite{azulay2019deep,bouchacourt2021grounding,engstrom2019exploring}.

\begin{figure}[t!]
\centering
\includegraphics[width=0.5\columnwidth]{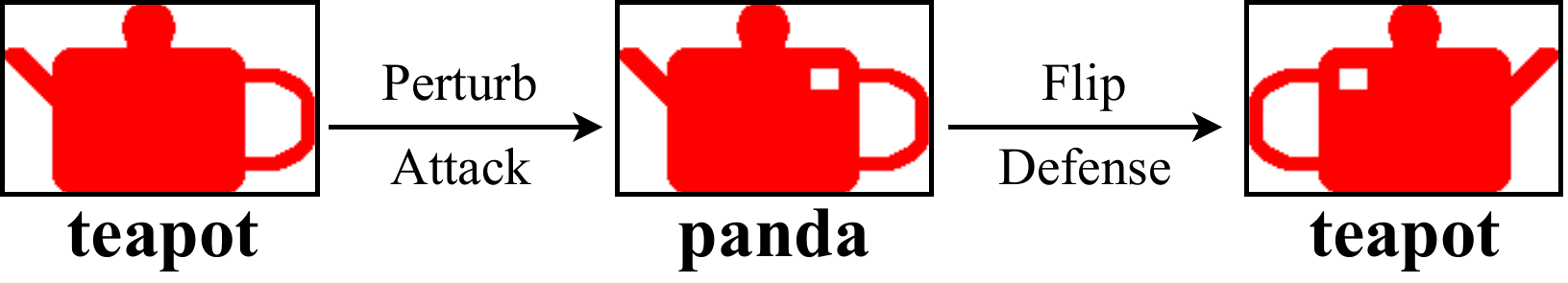}
\caption{
The flip symmetry defense against zero-knowledge adversaries reverts adversarial images to their correct classification by horizontally flipping the images before classification. The defense classifies non-adversarial images in the same way.
}
\label{fig:fig1}
\end{figure}

Against adversarial perturbation attacks causing misclassification, the CNN inability to handle symmetry well can be beneficial. Although an adversarial image classifies with a wrong label, a symmetric adversarial image generated by applying a symmetry to an adversarial image can classify with a label that is different from the wrong label of the adversarial image. Aiming to classify adversarial images correctly, we ask:
\\\\
\emph{Can we use the CNN inability to handle symmetry correctly for a defense that provides robustness against adversarial perturbation attacks?}
\\

Addressing this question, we design a novel symmetry defense that only uses symmetry to counter adversarial perturbation attacks. The proposed symmetry defense makes the following main contributions:

\begin{itemize}
  
  \item We show that the proposed symmetry defense succeeds against gradient-based attacks and a random search attack without using adversarial images or attack knowledge. In contrast, the current best defense needs attack knowledge to train the classifier with adversarial images.

  \item The symmetry defense counters zero-knowledge adversaries with near-default accuracies by using either the horizontal flip symmetry or an artificial pixel inversion symmetry. Results are shown in Table~\ref{table:zero_flip} and in Table~\ref{table:zero_invert}.

  \item The defense also counters perfect-knowledge adversaries with near-default accuracies, as shown in Table~\ref{table:perfect_both_jitter}. Against such adversaries, the defense uses a symmetry subgroup that consists of the identity symmetry, the mirror symmetry (also called horizontal flip), the pixel inversion symmetry, and the symmetry that combines the mirror flip and the pixel inversion symmetry.

  \item The defense counters adaptive attacks that use symmetry against the defense because an attack loss function applied to symmetric images depends on the function value of symmetric images, that is, on the CNN output evaluated at these images. Loss functions measure the distance between the function output and a label value. Since the function output can be different for symmetric images due to the CNN inability to handle symmetry well, the optimization of adaptive attacks that incorporate symmetry in their loss functions is not optimal.


  \item The usage of the pixel intensity inversion symmetry, discussed in Section~\ref{sec:zero_invert} and in Section~\ref{sec:perfect}, that does not exist in natural images of the dataset means that the proposed defense could be applied even to datasets without existing symmetries.
  
  \item The symmetry defense maintains and even exceeds the non-adversarial accuracy against perfect-knowledge adversaries, as shown in Table~\ref{table:perfect_both_jitter}.
\end{itemize}

\section{Related Work and Background}\label{sec:rel_work}

\subsection{Symmetry, Equivariance and Invariance in CNNs.} \label{sec:equi_inv}

\textbf{Symmetry} of an object is a transformation that leaves that object invariant. Image symmetries include rotation, horizontal flipping, and inversion~\cite{miller1973symmetry}. We provide definitions related to symmetry groups in Appendix~1.
A function $f$ is \textbf{equivariant} with respect to a transformation $\mathcal{T}$ if they commute with each-other~\cite{schmidt2012learning}: $f \circ T = T \circ f$. 
\textbf{Invariance} is a special case of equivariance where the $T$ transformation applied after the function is the identity transformation~\cite{schmidt2012learning}: $f \circ T = f$.

CNNs stack equivariant convolution and pooling layers~\cite{goodfellow2009measuring} followed by an invariant map in order to learn invariant functions~\cite{bronstein2021geometric} with respect to symmetries, following a standard blueprint used in machine learning~\cite{bronstein2021geometric,higgins2022symmetry}.
Translation invariance for image classification means that the position of an object in an image should not impact its classification.
To achieve translation invariance, CNN convolutional layers~\cite{krizhevsky2012imagenet,lecun1995convolutional} compute feature maps over the translation symmetry group~\cite{gens2014deep,sokolic2017generalization} using kernel sliding~\cite{gens2014deep,lecun1989backpropagation}. CNN pooling layers enable local translation invariance~\cite{bronstein2021geometric,dieleman2016exploiting,goodfellow2009measuring}. The pooling layers of CNNs positioned after convolutional layers enable local invariance to translation~\cite{dieleman2016exploiting} because the output of the pooling operation does not change when the position of features changes within the pooling region. Cohen and Welling~\cite{cohen2016group} show that convolutional layers, pooling, arbitrary pointwise nonlinearities, batch normalization, and residual blocks are equivariant to translation. CNNs learn invariance with respect to symmetries such as rotations, horizontal flipping, and scaling with data augmentation, which adds to the training dataset images obtained by applying symmetries to original images~\cite{krizhevsky2012imagenet}. For ImageNet, data augmentation can consist of a random crop, horizontal flip, color jitter, and color transforms of original images~\cite{madrylabrobustness}.

\textbf{CNN Lack of Translation Equivariance.} Studies suggest that CNNs are not equivariant to translation in CNNs~\cite{azulay2019deep,bouchacourt2021grounding,engstrom2019exploring,kayhan2020translation,zhang2019making}, not even to small translations or rotations~\cite{engstrom2019exploring}. Bouchacourt et al.~\cite{bouchacourt2021grounding} claim that the CNN translation invariance is approximate and that translation invariance is primarily learned from the data and data augmentation. The cause of translation invariance has been attributed to aliasing effects caused by the subsampling of the convolutional stride~\cite{azulay2019deep}, by max pooling, average pooling, and strides~\cite{zhang2019making}, or by image boundary effects~\cite{kayhan2020translation}.

\textbf{CNN Data Augmentation Marginally Effective.} Studies show that data augmentation is only marginally effective~\cite{cohen2016group,azulay2019deep,kohler2020equivariant,bouchacourt2021grounding,engstrom2019exploring} at incorporating symmetries because CNNs cannot learn invariances with data augmentation~\cite{azulay2019deep,bouchacourt2021grounding,engstrom2019exploring}. Engstrom et al.~\cite{engstrom2019exploring} find that data augmentation only marginally improves invariance. Azulay and Weiss~\cite{azulay2019deep} find that data augmentation only enables invariance to symmetries of images that resemble dataset images. Bouchacourt et al.~\cite{bouchacourt2021grounding} claim that non-translation invariance is learned from the data independently of data augmentation.

\textbf{Other Equivariance CNNs Approaches Have Dataset Limitation.} CNN architectures that handle symmetry better have only been shown to work for simple MNIST~\cite{lecun1998mnist} or CIFAR10~\cite{krizhevsky2009learning} or synthetic datasets, not ImageNet
\cite{schmidt2012learning,bruna2013invariant,sifre2013rotation,gens2014deep,cohen2016group,dieleman2016exploiting,zhou2017oriented,marcos2017rotation,finzi2020generalizing,romero2020group}.

\subsection{Adversarial Perturbation Attacks}

Szegedy et al.~\cite{szegedy2013intriguing} defined the problem of generating adversarial images as starting from original images and adding a small perturbation that results in misclassification. Szegedy et al.~\cite{szegedy2013intriguing} formalized the generation of adversarial images as a minimization of the sum of perturbation and an adversarial loss function, as shown in Appendix~2. The loss function uses the distance between obtained function output values and desired function output values.

Most attacks use the classifier gradient to generate adversarial perturbation~\cite{carlini2017towards,madry2017towards}, but random search~\cite{andriushchenko2020square} is also used. 

\textbf{PGD Attack.} PGD is an iterative white-box attack with a parameter that defines the magnitude of the perturbation of each step. PGD starts from an initial sample point $x_0$ and then iteratively finds the perturbation of each step and projects the perturbation on an $L_p$-ball.

\textbf{Auto-PGD Attack.} Auto-PGD (APGD)~\cite{croce2020reliable} is a variant of PGD that varies the step size and can use two different loss functions to achieve a stronger attack.

\textbf{Square Attack.} The Square Attack~\cite{andriushchenko2020square} is a score-based, black-box, random-search attack based on local randomized square-shaped updates.

\textbf{Fast Adaptive Boundary} The white-box Fast Adaptive Boundary attack (FAB)~\cite{croce2020minimally} aims to find the minimum perturbation needed to change the classification of an original sample. However, FAB does not scale to ImageNet because of the large number of dataset classes.

\textbf{AutoAttack.} AutoAttack~\cite{croce2020reliable} is a parameter-free ensemble of attacks that includes: APGD\textsubscript{CE} and APGD\textsubscript{DLR}, FAB~\cite{croce2020minimally} and Square Attack~\cite{andriushchenko2020square}.

\subsection{Adversarial Defenses}

\textbf{Adversarial Training.} AT~\cite{kurakin2016adversarial,madry2017towards,szegedy2013intriguing} trains classifiers with correctly-labeled adversarial images and is one of the first and few defenses that have not been defeated. The robust PGD AT defense~\cite{madry2017towards} is formulated as a robust optimization problem and is considered one of the most successful adversarial defenses~\cite{lindqvist2022novel}. AT usage of adversarial images during training increases training time and makes AT reliant on attack knowledge, which is unrealistic for actual attacks.

\textbf{Failed Defenses.} Many other defenses have been shown to fail against an adaptive adversary. For example, defensive distillation is not robust to adversarial attacks~\cite{carlini2016defensive}, many adversarial detection defenses have been bypassed~\cite{carlini2017adversarial,carlini2017magnet}, and obfuscated gradient defenses~\cite{athalye2018obfuscated} and other defenses have been circumvented~\cite{tramer2020adaptive}.

\subsection{Summary of Related Work}

Relevant to the proposed defense, we derive the following key points from the related work:
\begin{itemize} [leftmargin=*]
  \setlength\itemsep{0em}
    \item CNNs do not achieve full equivariance with respect to symmetries such as translation, horizontal flipping, and rotation despite being designed and trained with data augmentation to incorporate these symmetries.
  \item Adversarial perturbation attacks are still an open problem as the best current AT defense requires knowledge of the attack.
  \item Approaches to better incorporate symmetries into CNNs have yet to succeed for big datasets such as ImageNet.
\end{itemize}

\section{Threat Model}\label{sec:threat}

Based on recommendations for evaluating adversarial defenses~\cite{carlini2019evaluating}, the threat model consists of three cases:

\begin{itemize} [leftmargin=*]
  \setlength\itemsep{0em}
  \item \textbf{Zero-Knowledge Adversary.} The adversary is unaware of the symmetry defense.
  \item \textbf{Perfect-Knowledge Adversary.} The adversary is aware of the symmetry defense and adapts the attack.
  \item \textbf{Limited-Knowledge Adversary.} Based on~\cite{carlini2019evaluating}, this threat only needs to be evaluated if the zero-knowledge attack fails and the perfect-knowledge attack succeeds. Since the defense succeeds against both zero-knowledge and perfect-knowledge adversaries, we do not evaluate this case.
\end{itemize}

\section{The Proposed Symmetry Defense}

The inability of CNNs to handle symmetries well causes boundaries in symmetric settings to differ. Following, we note \underline{$f$} as the CNN function that finds the classifier boundary and \underline{$T$} as the pixel intensity inversion symmetry.

\subsection{CNN classifier boundary function $f$ lacks equivariance with respect to symmetries}

As shown in Section~\ref{sec:rel_work}, CNNs can classify symmetric images differently. This lack of invariance of CNN classification with respect to symmetries~\cite{azulay2019deep,bouchacourt2021grounding,engstrom2019exploring,kayhan2020translation} indicates that CNN function $f$ of finding the classification boundary lacks equivariance with respect to symmetries. That is, CNN classifier boundaries near symmetric images are not symmetric. Otherwise, if the CNN classifier boundaries were symmetric around symmetric images, then symmetric images would classify the same, and CNN classification would not lack invariance.


Formally, we base our reasoning on the definition of equivariance as $f \circ T = T \circ f$, where $f$ is the function that finds the classifier boundary and $T$ is the symmetry transformation. For the CNN function $f$ that finds the classifier boundary, equivariance with respect to $T$ would mean that the CNN classifier boundary would be the same whether we apply $f$ or $T$ first. However, the lack of invariance of CNN classification with respect to symmetries~\cite{azulay2019deep,bouchacourt2021grounding,engstrom2019exploring,kayhan2020translation} indicates that $f \circ T \ne T \circ f$. Otherwise, if $f \circ T = T \circ f$, then the boundaries would be the same, and CNN classification would not lack invariance. Therefore, we conclude that function $f$ is not equivariant with respect to symmetries. Taking pixel inversion as the example $T$ symmetry, Figure~\ref{fig:fig1} shows that inverting all dataset images and finding the classifier boundary do not commute and would result in different class boundaries.

\begin{figure}[t!]
\centering
\includegraphics[width=0.55\columnwidth]{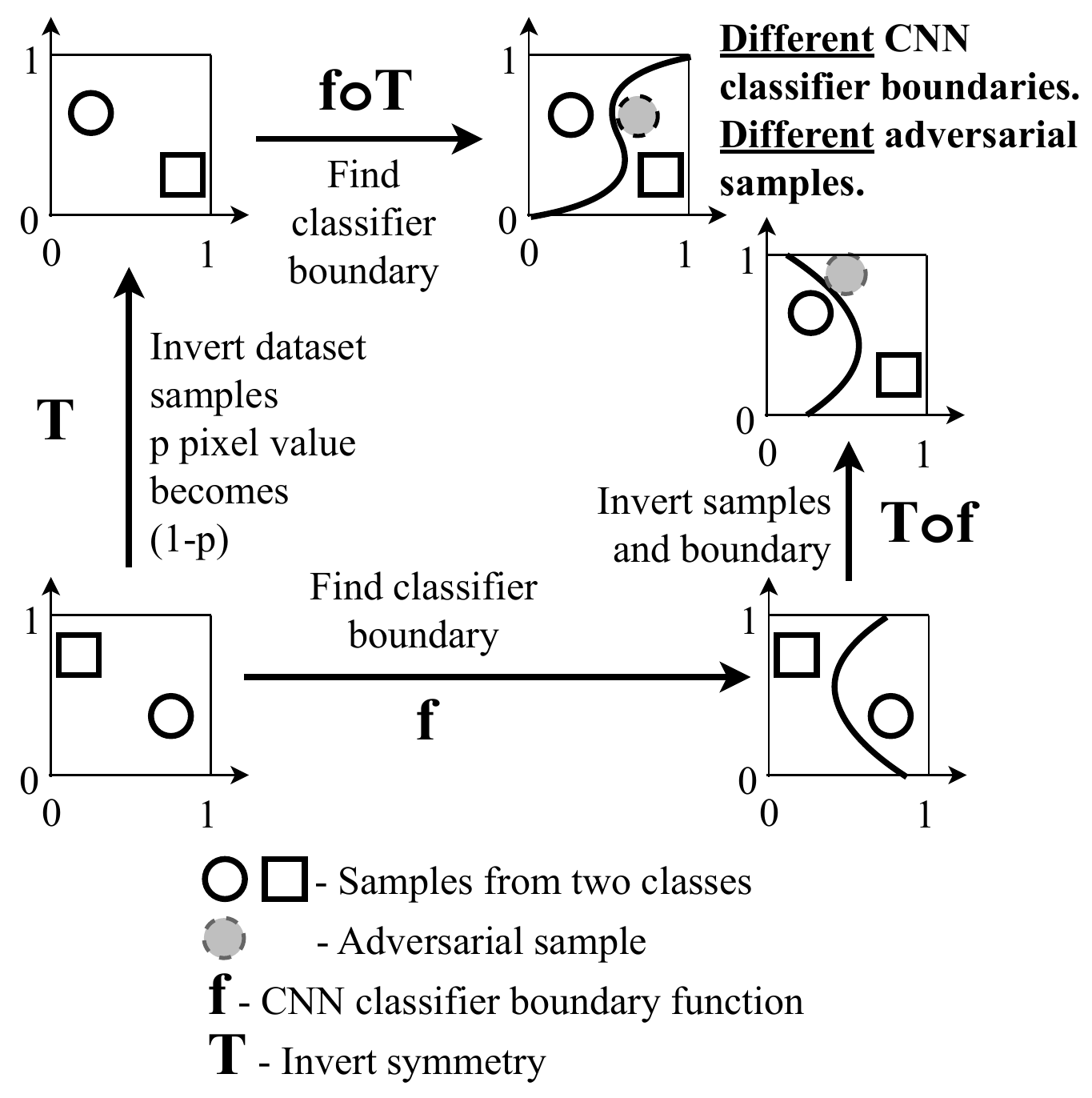}
\caption{
We show in two-dimensional space the lack of equivariance of the CNN classifier boundary function $f$ with respect to the pixel invert symmetry $T$. Applying first the inverse symmetry and then computing the classifier boundary function (up and right) produces a different boundary from computing the classifier function first and then applying the inverse symmetry (right and up). Due to the different class boundaries, adversarial images are also different.
}
\label{fig:fig1}
\end{figure}

\subsection{Adversarial Images Also Lack Equivariance}

As a result of the lack of equivariance in finding classifier boundaries with respect to symmetries, finding adversarial also lacks equivariance with respect to symmetries. In other words, adversarial images generated from symmetric original images do not correspond.
Conceptually, adversarial perturbation attacks~\cite{goodfellow6572explaining,szegedy2013intriguing,moosavi2016deepfool,madry2017towards,carlini2017towards} aim to change an original sample with a small perturbation in order to obtain an adversarial sample that is on the other side of the CNN classifier boundary and misclassifies as a result. Therefore, different class boundaries in symmetric settings would cause an adversarial perturbation attack to find adversarial images in symmetric settings that are different, as Figure~\ref{fig:fig1} shows.

\subsection{Adaptive Attacks Affected by CNN Inability to Handle Symmetry}

CNN inability to handle symmetry well means that the CNN function values for symmetric images can be different from the correct output. These different function values affect the loss function values that guide the optimization of attacks because loss functions depend on function output values. If the CNN outputs of symmetric images differ, the attack optimization can become non-optimal and steered to obtain adversarial images that classify correctly.

\section{Experimental Setting}

We evaluate the proposed symmetry defense in the threat model cases in Section~\ref{sec:threat}, based on~\cite{carlini2019evaluating}.
Our implementation is based on the robustness AT package implementation~\cite{madrylabrobustness} for ImageNet~\cite{russakovsky2015imagenet} with the same ResNet50~\cite{he2016deep} architecture and parameters.
ImageNet~\cite{russakovsky2015imagenet} is a $1000$-class dataset of over $1.2M$ training images and $50K$ testing images.
The ResNet50~\cite{he2016deep} architecture model is trained with the stochastic gradient descent (SGD) optimizer with a momentum of $0.9$, a learning rate decaying by a factor of $10$ every $50$ epochs, and a batch size of $256$. The classifier takes as input images with $[0,1]$ pixel value ranges. Based on~\cite{madrylabrobustness}, the evaluation is done on logits, non-softmax output.



\textbf{Default Data Augmentation.} We use the same data augmentation as in~\cite{madrylabrobustness} in the models we train: random resized crops, random horizontal flips, color jitter, and Fancy Principal Component Analysis (Fancy PCA)~\cite{krizhevsky2012imagenet}. The ColorJitter transform randomly changes the brightness, contrast, and saturation. Fancy PCA~\cite{krizhevsky2012imagenet} is a form of data augmentation that changes the intensities of RGB channels in training images by performing PCA analysis on ImageNet~\cite{deng2009imagenet} images and adding to every image multitudes of the principal components. The magnitudes are proportional to the eigenvalues and a random variable drawn from a Gaussian distribution with $0$ mean and $0.05$ standard deviation in~\cite{madrylabrobustness}.

\textbf{PGD Attacks.} We evaluate the proposed defense against $L_2$ and $L_\infty$ PGD~\cite{kurakin2016adversarial} attacks parameterized according to~\cite{madrylabrobustness} for ImageNet with $\epsilon$ values of $0.5$, $1.0$, $2.0$, $3.0$ for $L_2$ attacks, and $\epsilon$ values of $4/255$, $8/255$, $16/255$ for $L_\infty$ attacks. All PGD~\cite{kurakin2016adversarial} attacks have $100$ steps, with a step perturbation value defined as the ratio of $2.5 \times \epsilon$ over the number of steps, following~\cite{madrylabrobustness}. All PGD attacks are targeted according to~\cite{madrylabrobustness}, with the target label chosen uniformly at random among the labels other than the ground truth label.

\textbf{AutoAttack attacks.} We evaluate against APGD, and SquareAttack attacks with $1,000$ random images for each experiment based on~\cite{croce2020reliable} experiments with ImageNet. We do not evaluate against FAB because it does not scale to ImageNet due to the large number of ImageNet classes~\cite{croce2020reliable}. All APGD attacks are targeted. Square Attack~\cite{andriushchenko2020square} is not targeted based on~\cite{andriushchenko2020square}, with $10,000$ queries, $p = 0.02$ for $L_2$, and $p = 0.01$ for $L_{\infty}$. Square Attack~\cite{andriushchenko2020square} settings are based on settings for ImageNet in~\cite{andriushchenko2020square}. APGD\textsubscript{CE}~\cite{croce2020reliable} and APGD\textsubscript{DLR}~\cite{croce2020reliable} settings were based on attack settings for ImageNet in~\cite{croce2020reliable}.

\textbf{Tools.} The defense was written using PyTorch~\cite{NEURIPS2019_9015}. PGD attacks were generated with the Robustness (Python Library)~\cite{madrylabrobustness}, AutoAttack attacks were generated with the IBM Adversarial Robustness 360 Toolbox~(ART)~\cite{art2018}.

\subsection{Symmetry Defense Against Zero-Knowledge Adversaries}\label{sec:zero}

We assume that zero-knowledge adversaries know the model and its parameters but are unaware of the symmetry defense. We discuss the flip symmetry defense in Section~\ref{sec:zero_flip} and the intensity inversion symmetry defense in Section~\ref{sec:zero_invert}.

\subsubsection{Horizontal Flipping Symmetry Defense Against Zero-Knowledge Adversaries} \label{sec:zero_flip}

\begin{figure}[h!]
\centering
\includegraphics[width=0.7\columnwidth]{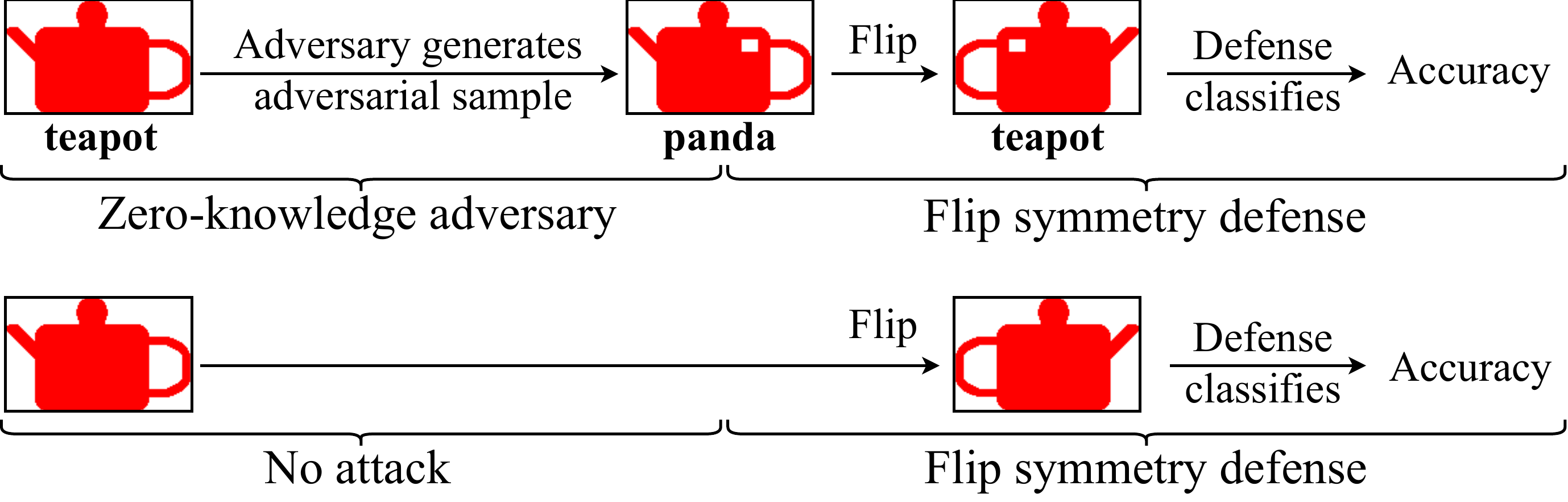}
\caption{
The flip symmetry defense against zero-knowledge adversaries horizontally flips images before classifying them. The defense uses the same model that the adversary uses to generate adversarial images.
}
\label{fig:flip}
\end{figure}

Both the adversary and the defense use a model trained with the default training dataset because horizontal flips are used in the default data augmentation.
Table~\ref{table:zero_flip} shows the experimental results of the defense using horizontal flip symmetry to counter zero-knowledge adversaries.
Figure~\ref{fig:flip} shows that the defense classifies an image by first horizontally flipping it and then classifying it with the same model used by the adversary to generate the adversarial images.

\begin{table}[h!]
\caption{Evaluation of the flip symmetry defense against zero-knowledge attacks.}
\centering
\setlength{\tabcolsep}{5pt}
\begin{tabular}{ll|rr}
\toprule
Norm & \multicolumn{1}{l|}{Attack} & No defense & \textbf{Proposed defense} \\
\hline
 & $\epsilon=0.0$  & 77.26\% & 77.15\% \\
\hline
\multirow{7}{*}{$L_2$}
& PGD - $\epsilon=0.5$    & 34.27\% & 76.13\% \\
& PGD - $\epsilon=1.0$    &  4.32\% & 75.21\% \\
& PGD - $\epsilon=2.0$    &  0.19\% & 74.25\% \\
& PGD - $\epsilon=3.0$    &  0.03\% & 73.74\% \\
& \textbf{*}APGD\textsubscript{CE} - $\epsilon=3.0$  &  0.0\%  & 75.0\% \\
& \textbf{*}APGD\textsubscript{DLR} - $\epsilon=3.0$ & 37.9\%  & 79.2\% \\
& \textbf{*}Square Attack - $\epsilon=5.0$           & 40.0\%  & 71.9\% \\
\hline
\multirow{6}{*}{$L_\infty$}
& PGD - $\epsilon=4/255$  &  0.00\% & 74.27\% \\
& PGD - $\epsilon=8/255$  &  0.00\% & 73.24\% \\
& PGD - $\epsilon=16/255$ &  0.00\% & 70.13\% \\
& \textbf{*}APGD\textsubscript{CE} - $\epsilon=4/255$  & 0.0\% & 71.8\% \\
& \textbf{*}APGD\textsubscript{DLR} - $\epsilon=4/255$ & 0.0\% & 78.7\% \\
& \textbf{*}Square Attack - $\epsilon=0.05$            & 4.9\% & 47.5\% \\
\bottomrule
\multicolumn{4}{p{340pt}}{\textbf{*}Evaluation based on $1,000$ random images.}\\
\multicolumn{4}{p{340pt}}{
The defense achieves close to the non-adversarial accuracy against most attacks, even exceeding it for APGD\textsubscript{DLR}. Furthermore, the defense also exceeds the performance of the robust PGD AT defense~\cite{madrylabrobustness} against PGD.
The flip symmetry defense against a zero-knowledge adversary maintains the default non-adversarial accuracy.
}\\
\end{tabular}
\label{table:zero_flip}
\end{table}

\subsubsection{Inversion Symmetry Defense Against Zero-Knowledge Adversaries} \label{sec:zero_invert}

\begin{figure}[h!]
\centering
\includegraphics[width=0.7\columnwidth]{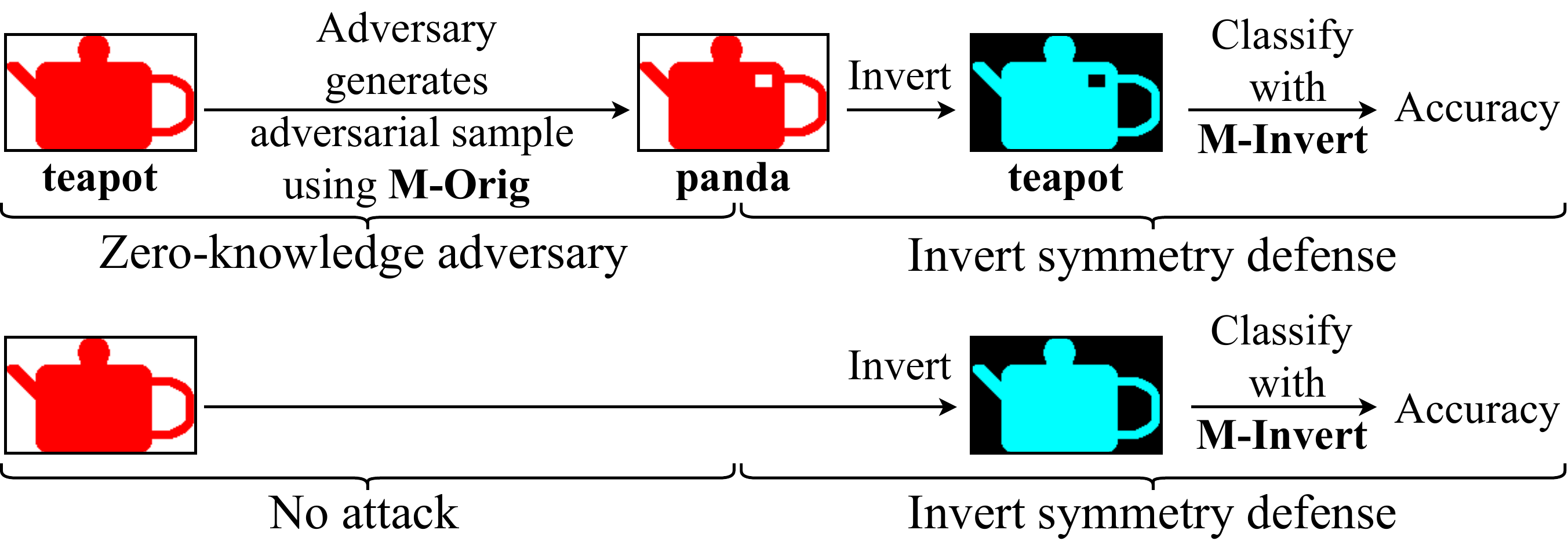}
\caption{
Whether an image is original or an adversarial image generated with the M-Orig model by attacks, the defense inverts the image and classifies the inverted image with the M-Invert model.
}
\label{fig:invert}
\end{figure}

The intensity inversion symmetry changes image pixel values from $p \in [0,1]$ to $1-p$ and is not present in the natural images. We train two models with the same preprocessing, parameters, and model architecture: M-Orig with original images and M-Invert with inverted images. The adversary generates adversarial images using the M-Orig model. The defense classifies a sample by inverting it and then classifying it with the M-Invert model. Figure~\ref{fig:invert} outlines the invert symmetry defense, and Table~\ref{table:zero_invert} shows that the inversion symmetry defense achieves near-default accuracies against most attacks and maintains non-adversarial accuracy.

\begin{table}[h!]
\caption{Evaluation of the invert symmetry defense against zero-knowledge attacks.}
\centering
\setlength{\tabcolsep}{5pt}
\begin{tabular}{ll|rr}
\toprule
Norm & \multicolumn{1}{l|}{Attack} & No defense & \textbf{Proposed defense} \\
\hline
 & $\epsilon=0.0$  & 77.26\% & 76.88\% \\
\hline
\multirow{7}{*}{$L_2$}
& PGD - $\epsilon=0.5$    & 34.27\% & 75.87\% \\
& PGD - $\epsilon=1.0$    &  4.32\% & 75.10\% \\
& PGD - $\epsilon=2.0$    &  0.19\% & 74.33\% \\
& PGD - $\epsilon=3.0$    &  0.03\% & 74.02\% \\
& \textbf{*}APGD\textsubscript{CE} - $\epsilon=3.0$  &  0.0\% & 73.8\% \\
& \textbf{*}APGD\textsubscript{DLR} - $\epsilon=3.0$ & 34.6\% & 76.0\% \\
& \textbf{*}Square Attack - $\epsilon=5.0$           & 40.4\% & 72.6\% \\
\hline
\multirow{6}{*}{$L_\infty$}
& PGD - $\epsilon=4/255$  & 0.00\% & 74.54\% \\
& PGD - $\epsilon=8/255$  & 0.00\% & 73.83\% \\
& PGD - $\epsilon=16/255$ & 0.00\% & 72.08\% \\
& \textbf{*}APGD\textsubscript{CE} - $\epsilon=4/255$  & 0.0\% & 69.8\% \\
& \textbf{*}APGD\textsubscript{DLR} - $\epsilon=4/255$ & 0.2\% & 71.2\% \\
& \textbf{*}Square Attack - $\epsilon=0.05$            & 4.0\% & 48.1\% \\
\bottomrule
\multicolumn{4}{p{340pt}}{\textbf{*}Evaluation based on $1,000$ random images.}\\
\multicolumn{4}{p{340pt}}{
Similarly to the flip symmetry defense, the invert symmetry defense achieves near-default accuracy against most attacks, exceeding the performance of the robust PGD AT defense~\cite{madrylabrobustness} for PGD attacks. In addition, the defense accuracy maintains the accuracy for non-adversarial images.
}\\
\end{tabular}
\label{table:zero_invert}
\end{table}

\subsection{Symmetry Defense Against Perfect-Knowledge Adversaries} \label{sec:perfect}

Here, we assume that the adversary is aware of the proposed symmetry defense and can adapt the attack to the defense.

\textbf{The defense needs more than one symmetry.} To counter a perfect-knowledge adversary, the defense must use more than one symmetry transformation. An adversary that is aware of the defense could apply the flip or invert symmetry after generating the adversarial sample, which would cancel out the symmetry transformation applied by the defense in Section~\ref{sec:zero} (flipping or inverting an image twice reverts it to the same image). In addition, the defense against a perfect-knowledge adversary would need to use such symmetry transformations that their possible combinations are reasonably limited in number to enable the defense to conduct experiments for all cases.



\textbf{Definition of the discrete subgroup of transformations.} We define subgroup $H$ with a discrete set of transformations $H=\{e,a,b,c\}$, where $e,a,b,c$ denote the identity, horizontal flipping, intensity inversion, the composition of flipping and inversion. The operation $*$ means that one transformation follows another. The Cayley Table~\ref{table:cayley} shows that the subgroup is closed since compositions of the elements also belong to the subgroup. The defined $H$ subgroup is known as the \textbf{Klein four-group}. In this four-element group, each element is its own inverse, and composing any two non-identity elements results in the third non-identity element. Another way to define the Klein four-group $H$ is: $H=\{a,b|a^2=b^2=(a*b)^2=e\}$.

\begin{table}[h!]
\caption{The Cayley table shows that the defined $H$ subgroup is closed because all compositions of symmetries belong to the subgroup.}
\centering
\setlength{\tabcolsep}{3pt}
\begin{tabular}{rcc|c|c|c|c|}
\toprule
&&    *               & $e$ & $a$ & $b$ & $c$ \\
\hline
identity& - &$e$        & $e$ & $a$ & $b$ & $c$ \\
\hline
flip& - &$a$            & $a$ & $e$ & $c$ & $b$ \\
\hline
invert& - &$b$          & $b$ & $c$ & $e$ & $a$ \\
\hline
flip and invert& - &$c$ & $c$ & $b$ & $a$ & $e$ \\
\bottomrule
\end{tabular}
\label{table:cayley}
\end{table}

\begin{theorem}
$H=\{e,a,b,c\}$ is a subgroup of the group of symmetry transformations of images.
\end{theorem}

\begin{proof}
Based on the finite subgroup criterion in Section~\ref{sec:rel_work}, a finite subset of a group should need only to be nonempty and closed under operation $*$~\cite{dummit2004abstract}. $H$ is nonempty because it has four elements and is a subset of the symmetry transformations of images. Based on the definition of closure in Section~\ref{sec:rel_work}, for $H$ to be closed under the $*$ operation, we need to show that $\forall a,b \in H$, we get that $a*b^{-1} \in H$. Table~\ref{table:cayley} shows that $\forall a,b \in H$, we get that $a*b \in H$. Table~\ref{table:cayley} also shows that each element is its own inverse element because $\forall b \in H$, we get $b*b=e$, which means that $b=b^{-1}$. From $\forall a,b \in H$, $a*b \in H$ and $\forall b \in H$, $b=b^{-1}$, we derive that $\forall a,b \in H$, $a*b^{-1} \in H$.
\end{proof}

\begin{figure}[h!]
\centering
\includegraphics[width=0.5\columnwidth]{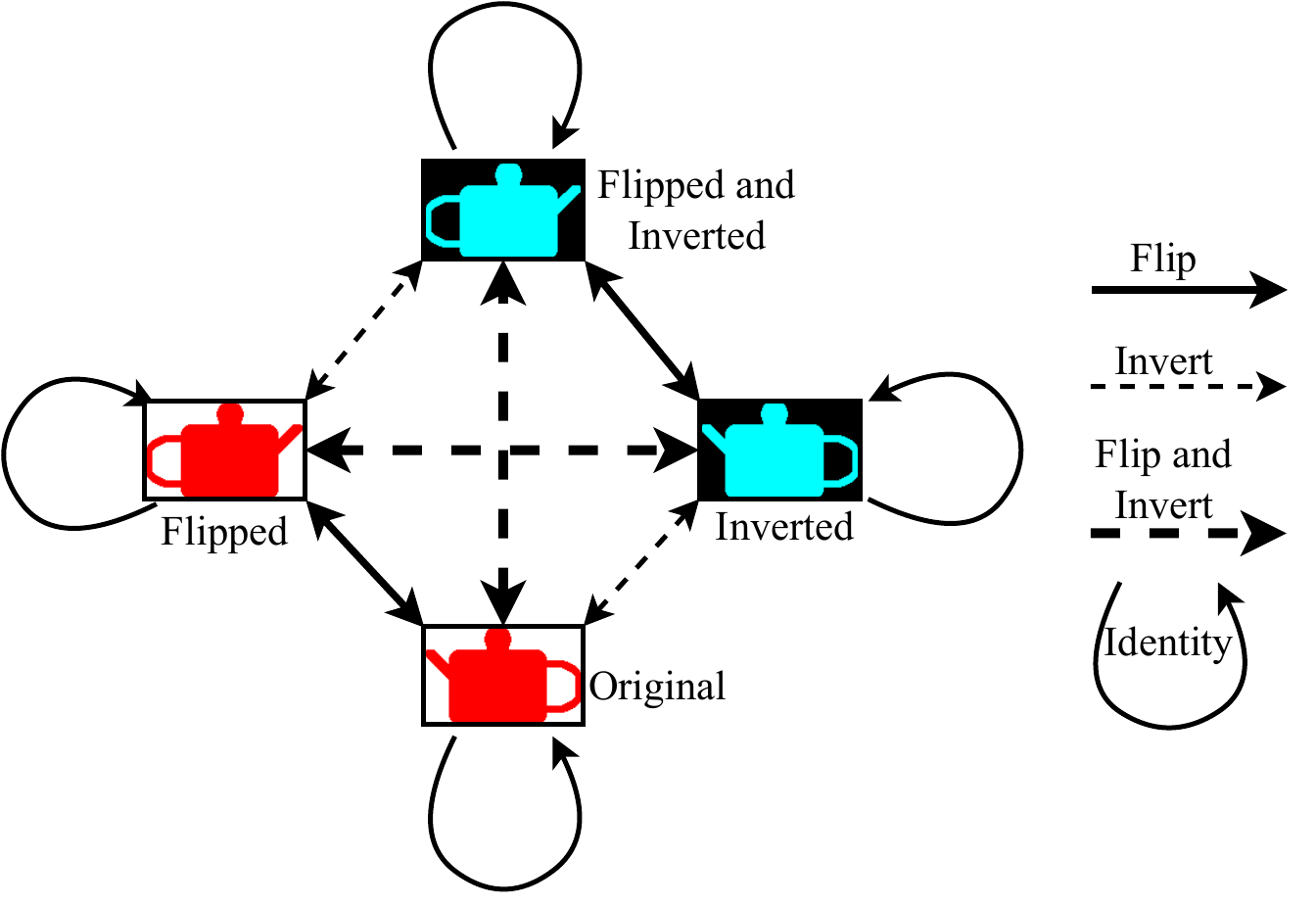}
\caption{No matter what sequence of $H$ symmetries is performed, an image will be in one of the four states shown.}
\label{fig:group}
\end{figure}

\begin{figure*}[h!]
\centering
\includegraphics[width=1.0\columnwidth]{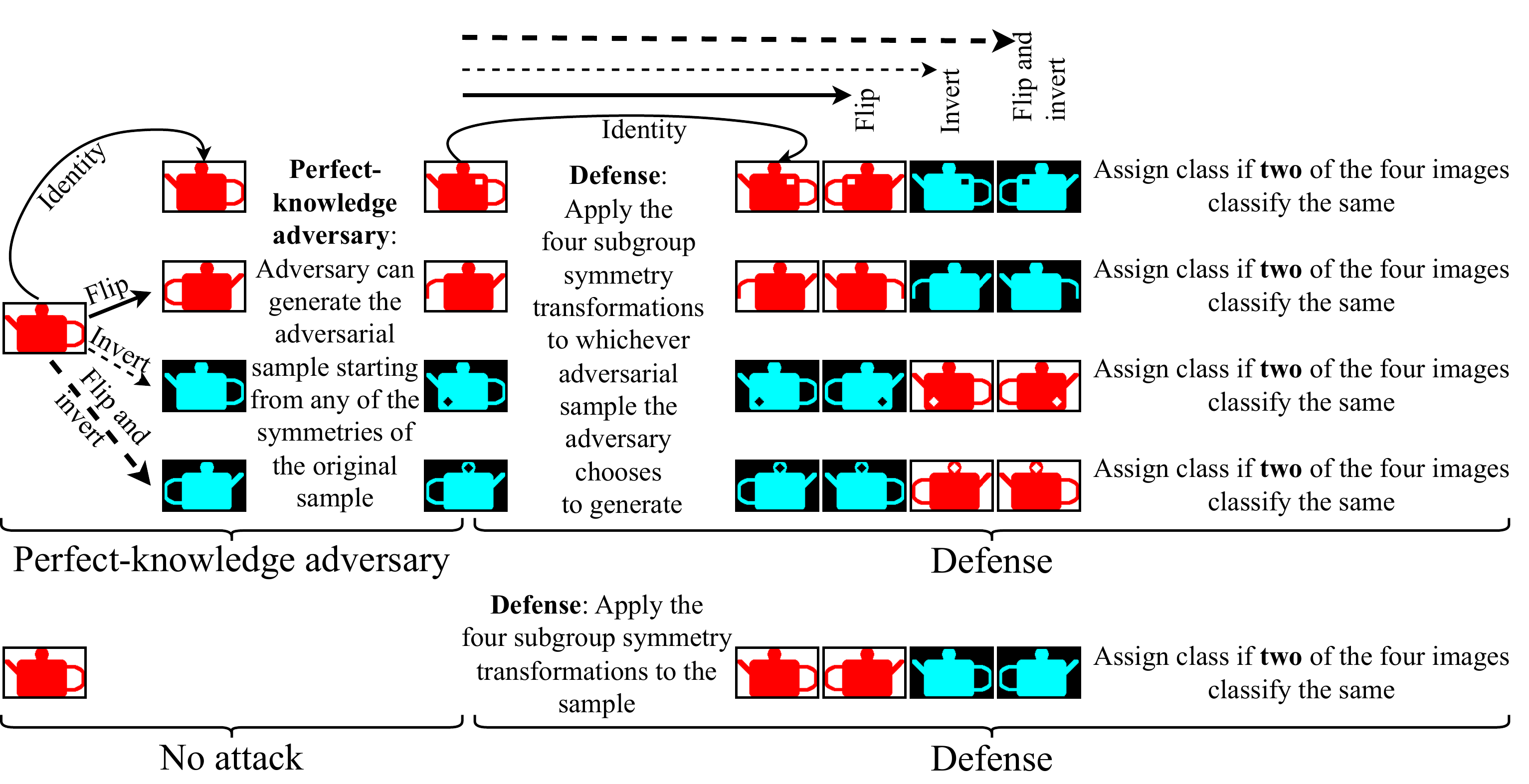}
\caption{The symmetry subgroup defense against adaptive perfect-knowledge adversaries considers that symmetries could be applied to images by adaptive adversaries both before and after adversarial generation. Moreover, even if adversaries applied consecutive subgroup symmetries to images, they would not result in any other cases due to the closure property of the symmetry subgroup.
}
\label{fig:fig-perfect-adv}
\end{figure*}

Figure~\ref{fig:group} shows that the defined subgroup $H$ confines the states that an image can be in after any consecutive combination of symmetries from $H$, which facilitates evaluating all the possible combinations of $H$ transformations that an adversary can apply before or after the adversarial generation, as shown in Figure~\ref{fig:fig-perfect-adv}.

\textbf{Training.} We impose the $H$ subgroup shown in Figure~\ref{fig:group} on the CNN model by augmenting the original dataset with inverted images and using the default data augmentation, which includes horizontal flips.

\textbf{Evaluation.} The defense evaluates both original and adversarial images in the same way. To classify an image, the proposed defense applies all four $H$ subgroup symmetries to the image and classifies all four images with the same model used by the attack to generate attacks. The defense assigns a class to the image if the classification labels of two of the four symmetric images agree. Experimental results are shown in Table~\ref{table:perfect_both_jitter}. The defense achieves near-default accuracies for many attacks, surpassing the current best defense, robust PGD AT defense~\cite{madrylabrobustness}, against PGD attacks. The defense also exceeds the default accuracy.

\begin{table*}[hbt!]
\caption{Accuracy evaluation of the proposed symmetry defense against an adaptive perfect-knowledge adversary.}
\centering
\setlength{\tabcolsep}{3pt}
\begin{tabular}{ll|r|rrrr}
\toprule
\multicolumn{2}{c|}{} & \multicolumn{1}{c|}{No} & \multicolumn{4}{c}{\textbf{Proposed defense}} \\
Norm & \multicolumn{1}{l|}{Attack} & defense & \multicolumn{4}{c}{ Adversarial images generated from} \\
\cline{4-7} 
  
    &  &   & \multicolumn{1}{r}{ Original} & \multicolumn{1}{r}{ Flipped} & \multicolumn{1}{r}{ Inverted} & \multicolumn{1}{r}{ Flip. \& Inv.}\\
    
  &  &   & images & images & images & images \\
  
\hline
&$ \epsilon=0.0$  & 77.26\% & 78.30\% & 78.30\% & 78.30\% & 78.30\% \\
\hline
\multirow{7}{*}{$L_2$}

& PGD - $\epsilon=0.5$                        & 34.27\% & 75.65\% & 75.80\% & 75.72\% & 75.69\% \\
& PGD - $\epsilon=1$                          &  4.32\% & 73.16\% & 73.27\% & 73.46\% & 73.35\% \\
& PGD - $\epsilon=2$                          &  0.19\% & 70.80\% & 70.81\% & 70.80\% & 70.77\% \\
& PGD - $\epsilon=3$                          &  0.03\% & 69.75\% & 69.82\% & 69.83\% & 69.89\% \\
& \textbf{*}APGD\textsubscript{CE} - $\epsilon=3.0$    &  0.0\% & 70.5\% & 70.4\% & 71.0\% & 70.7\% \\
& \textbf{*}APGD\textsubscript{DLR} - $\epsilon=3.0$   & 38.5\% & 77.7\% & 77.7\% & 78.0\% & 77.8\% \\
& \textbf{*}SquareAttack - $\epsilon=5.0$              & 43.0\% & 74.3\% & 74.5\% & 73.5\% & 73.1\% \\
\hline
\multirow{6}{*}{$L_\infty$}
& PGD - $\epsilon=4/255$                      & 0.00\% & 70.68\% & 70.83\% & 70.77\% & 70.84\% \\
& PGD - $\epsilon=8/255$                      & 0.00\% & 69.08\% & 69.12\% & 69.16\% & 69.21\% \\
& PGD - $\epsilon=16/255$                     & 0.00\% & 65.26\% & 65.16\% & 65.32\% & 65.17\% \\
& \textbf{*}APGD\textsubscript{CE} - $\epsilon=4/255$  & 0.0\% & 66.5\% & 66.9\% & 68.1\% & 66.6\% \\
& \textbf{*}APGD\textsubscript{DLR} - $\epsilon=4/255$ & 0.6\% & 68.5\% & 68.1\% & 68.9\% & 69.4\% \\
& \textbf{*}SquareAttack - $\epsilon=0.05$             & 5.3\% & 49.0\% & 47.7\% & 48.7\% & 48.1\% \\
\bottomrule
\multicolumn{7}{p{320pt}}{ \textbf{*}Evaluation based on $1,000$ random images.}\\
\multicolumn{7}{p{320pt}}{
The symmetry defense against perfect-knowledge adversaries exceeds the default accuracy and achieves near-default accuracies for many attacks. The defense also surpasses the robust PGD AT defense~\cite{madrylabrobustness} against PGD attacks.
}
\end{tabular}
\label{table:perfect_both_jitter}
\end{table*}

\subsubsection{Adversary Adapts to the Defense}

We discuss how a perfect-knowledge adversary could adapt to the defense before, during, and after the generation of adversarial images.

\textbf{1) During the adversarial perturbation generation.} An adaptive adversarial attack must use symmetry in its adaptation because the symmetry defense makes no other changes. Both gradient-based attacks and the random-search SquareAttack use loss functions that depend on the CNN function evaluation. Adaptive attacks using symmetry to counter the defense would need to update their optimization with loss functions that use CNN function values evaluated at symmetric images. However, these CNN function values can be affected by CNNs' inability to handle symmetry correctly. Wrong function outputs for symmetric images will affect the loss function, making the optimizations of attacks non-optimal. We implement an adaptive PGD attack where the adversary attacks all four symmetries of the image by maximizing the sum of their losses, aiming to cause the misclassification of all symmetries of the image. We experiment with an $L_\infty$ norm of $16/255$, the strongest PGD attack in~\cite{madrylabrobustness}. The defense obtains an accuracy of \textbf{75.55\%} against the adaptive attack, exceeding the 65.16\% to 65.32\% accuracies in Table~\ref{table:perfect_both_jitter} obtained against the default PGD attack for the same norm and perturbation. Therefore, \textbf{the adaptive attack against the proposed defense fails}. The adaptive adversary cannot make any other adaptations based on non-symmetry changes because the preprocessing, parameters, and model architecture do not change from~\cite{madrylabrobustness}.





\textbf{2) Before the adversarial generation.} An adversary can apply a subgroup $H$ symmetry to the original image before the adversarial generation. Figure~\ref{fig:group} shows that even if the adversary applies any sequence of subgroup $H$ symmetries, the adversary can construct an adversarial image starting from either an original, a flipped, an inverted, or a flipped and inverted image. We evaluate all four cases in Table~\ref{table:perfect_both_jitter}.

\textbf{3) After the adversarial generation.} The adversary can apply any sequence of subgroup $H$ symmetries to the adversarial image after generating it. However, this would be irrelevant because the defense applies all four symmetries and would obtain the same images regardless of any sequence of subgroup symmetries that the adversary applies, based on Figure~\ref{fig:group}.

\subsection{Discussion of the Proposed Defense}

\textbf{Why adaptive attacks fail.} We explain the failure of adaptive attacks from two different viewpoints:

\begin{itemize}
  
  \item 1) Custom-loss adaptive attacks fail because their optimization is affected by the CNN inability to handle symmetry well. Such attacks need to use symmetry because the symmetry defense makes no other changes. As a result, these attacks would need to update their loss functions to use function values evaluated at symmetric images. However, CNN outputs for symmetric images can be different and incorrect, leading to loss function values that steer attack optimization to non-optimal adversarial images.
  
  \item 2) Adaptive attacks also fail because they are constrained in their optimizations by the perturbation value that attacks such as PGD, APGD, and SuareAttack take as input. These attacks search not in the entire space but for adversarial images with the given perturbation value. If the given perturbation value limits the attack search to where classifier boundaries are non-equivariant in symmetric settings, the attack will not be able to find an adversarial sample.

\end{itemize}


\textbf{Computational resources.} The proposed method has negligible computational overhead for the flip symmetry defense and roughly doubles the computational resources for the invert symmetry defense and the symmetry subgroup defense. Detailed computational analysis is in Appendix~3.

\textbf{Not a detection defense.} The proposed defense is not a detection defense because it classifies original and adversarial images in the same way, as shown in Figure~\ref{fig:flip}, Figure~\ref{fig:invert}, and Figure~\ref{fig:fig-perfect-adv}.

\textbf{Not a gradient obfuscation defense.} The defense does not rely on obfuscation because it keeps the exact preprocessing, parameters, and model as in~\cite{madrylabrobustness}.






\section{Conclusions}




The proposed symmetry defense succeeds with near-default accuracies against attacks ranging from being unaware of the defense to being aware of and adapting to it.
Importantly, the defense also defeats adaptive attacks that are aware of the symmetry defense.
The defense exceeds the classification accuracies of the current best defense, which uses adversarial images.
The defense's non-reliance on attack knowledge or adversarial images makes the defense applicable to realistic attack scenarios where the attack is unknown in advance.
The defense's preservation of classifier preprocessing, parameters, architecture, and training facilitates the deployment of the defense to classifiers.
The defense maintains the non-adversarial classification accuracy and even exceeds it against attacks aware of the defense.




\section*{Appendix 1: Definitions Related to Symmetry Groups}\label{sec:group_def} 

According to~\cite{dummit2004abstract}, a group is an ordered pair $(G,*)$ where $G$ is a set and $*$ is a binary operation on $G$ that satisfies these axioms:

\begin{itemize} [leftmargin=*]
  \setlength\itemsep{0em}
  \item \emph{Associativity.} $*$ is associative: $\forall a,b,c \in G$, $(a*b)*c=a*(b*c)$.
  \item \emph{Identity.} There exists an identity element $e \in G$, such that $a*e = e*a = a$, for $\forall a \in G$.
  \item \emph{Inverse.} Every element in G has an inverse: $\forall a \in G$, there exists $a^{-1} \in G$ such that $a*a^{-1} = a^{-1} *a = e$.
\end{itemize}

\textbf{Binary Operation.} According to~\cite{dummit2004abstract}, a binary operation $*$ on a set $G$ is a function $*$: $G \times G \mapsto G$. Instead of writing the binary operation $*$ on $a,b \in G$ as a function $*(a,b)$, we can write it as $a*b$.

\textbf{Closure.} Suppose that $*$ is a binary operation on the set G and $H$ is a subset of $G$. If $*$ is a binary operation on $H$, that is, $\forall a,b \in H$, $a*b \in H$, then $H$ is said to be closed under the $*$ binary operation~\cite{dummit2004abstract}.

\textbf{Group.} According to~\cite{dummit2004abstract}, a group is an ordered pair $(G,*)$ where $G$ is a set and $*$ is a binary operation on $G$ that satisfies the associativity, identity and inverse axioms.

\textbf{Subgroup.} According to~\cite{dummit2004abstract}, a subset $H$ of $G$ is a subgroup of $G$ if $H$ is nonempty and $H$ is closed under products and inverses (that is, $x,y \in H$ implies that $x^{-1} \in H$ and $x*y \in H$). A subgroup $H$ of group $G$ is written as $H \leq G$. Informally, the subgroup of a group $G$ is a subset of $G$, which is itself a group with respect to the binary operation defined in $G$.

\textbf{The Subgroup Criterion.} A subset $H$ of a group $G$ is a subgroup if and only if $H \neq \emptyset$ and $\forall x,y \in H$, $x*y^{-1} \in H$~\cite{dummit2004abstract}.

\textbf{The Finite Subgroup Criterion.} A finite subset $H$ is a subgroup if $H$ is nonempty and closed under $*$~\cite{dummit2004abstract}.

\section*{Appendix 2: The minimization of targeted adversarial perturbation attacks} \label{sec:minimization} 

The minimization for adversarial perturbation attacks targeted at a specific adversarial label was first formulated by Szegedy et al.~\cite{szegedy2013intriguing}:

\begin{align} \label{eq_targeted_attacks}
& {\text{minimize}}
& &  c \cdot \| \delta \| + L_{f}(x+\delta, l)  \\
& \text{such that} \nonumber
& &  x+\delta \in [0,1]^d,
\end{align}

where $f$ is the classifier function, $L_{f}$ is the classifier function loss, and $l$ is an adversarial label, $c$ is a constant, $\| \delta \|$ is the $L_p$ norm of perturbation.

\section*{Appendix 3: Computational Resources}\label{sec:resources} 
Here, we analyze the additional computational complexity of the proposed defense and the adversary.

\subsection*{Defense}

\textbf{Against a zero-knowledge adversary.} The flip symmetry defense uses the same computational complexity as a default classifier in training because it only trains one model with original images. The invert symmetry defense doubles the computational complexity of a default classifier in training because it trains two models with original and inverted images, respectively. In testing, there is $O(1)$ overhead per sample due to flipping or inverting the sample.

\textbf{Against a perfect-knowledge adversary.} The symmetry subgroup defense doubles the computational complexity of a default classifier in training because it trains one model with both original and inverted images. In testing, there is $O(1)$ overhead per sample due to flipping, inverting, or flipping and inverting the sample.

\subsection*{Adversary}

\textbf{Zero-knowledge adversary.} The zero-knowledge adversary is unaware of the defense and consumes the same resources as in the default case.

\textbf{Perfect-knowledge adversary.} The perfect-knowledge adversary can symmetrically transform the sample before and after generating the adversarial sample, using $O(1)$ additional computing resources per sample.

\bibliographystyle{splncs04}
\bibliography{egbib}

\begin{thebibliography}{10}
\providecommand{\url}[1]{\texttt{#1}}
\providecommand{\urlprefix}{URL }
\providecommand{\doi}[1]{https://doi.org/#1}

\bibitem{andriushchenko2020square}
Andriushchenko, M., Croce, F., Flammarion, N., Hein, M.: Square attack: a
  query-efficient black-box adversarial attack via random search. In: Computer
  Vision--ECCV 2020: 16th European Conference, Glasgow, UK, August 23--28,
  2020, Proceedings, Part XXIII. pp. 484--501. Springer (2020)

\bibitem{athalye2018obfuscated}
Athalye, A., Carlini, N., Wagner, D.: Obfuscated gradients give a false sense
  of security: Circumventing defenses to adversarial examples. arXiv preprint
  arXiv:1802.00420  (2018)

\bibitem{azulay2019deep}
Azulay, A., Weiss, Y.: Why do deep convolutional networks generalize so poorly
  to small image transformations? Journal of Machine Learning Research
  \textbf{20},  1--25 (2019)

\bibitem{bouchacourt2021grounding}
Bouchacourt, D., Ibrahim, M., Morcos, A.: Grounding inductive biases in natural
  images: invariance stems from variations in data. Advances in Neural
  Information Processing Systems  \textbf{34},  19566--19579 (2021)

\bibitem{bronstein2021geometric}
Bronstein, M.M., Bruna, J., Cohen, T., Veli{\v{c}}kovi{\'c}, P.: Geometric deep
  learning: Grids, groups, graphs, geodesics, and gauges. arXiv preprint
  arXiv:2104.13478  (2021)

\bibitem{bruna2013invariant}
Bruna, J., Mallat, S.: Invariant scattering convolution networks. IEEE
  transactions on pattern analysis and machine intelligence  \textbf{35}(8),
  1872--1886 (2013)

\bibitem{carlini2019evaluating}
Carlini, N., Athalye, A., Papernot, N., Brendel, W., Rauber, J., Tsipras, D.,
  Goodfellow, I., Madry, A., Kurakin, A.: On evaluating adversarial robustness.
  arXiv preprint arXiv:1902.06705  (2019)

\bibitem{carlini2016defensive}
Carlini, N., Wagner, D.: Defensive distillation is not robust to adversarial
  examples. arXiv preprint arXiv:1607.04311  (2016)

\bibitem{carlini2017adversarial}
Carlini, N., Wagner, D.: Adversarial examples are not easily detected:
  Bypassing ten detection methods. In: Proceedings of the 10th ACM Workshop on
  Artificial Intelligence and Security. pp. 3--14. ACM (2017)

\bibitem{carlini2017magnet}
Carlini, N., Wagner, D.: Magnet and" efficient defenses against adversarial
  attacks" are not robust to adversarial examples. arXiv preprint
  arXiv:1711.08478  (2017)

\bibitem{carlini2017towards}
Carlini, N., Wagner, D.: Towards evaluating the robustness of neural networks.
  In: 2017 IEEE Symposium on Security and Privacy (SP). pp. 39--57. IEEE (2017)

\bibitem{cohen2016group}
Cohen, T., Welling, M.: Group equivariant convolutional networks. In:
  International conference on machine learning. pp. 2990--2999. PMLR (2016)

\bibitem{croce2020minimally}
Croce, F., Hein, M.: Minimally distorted adversarial examples with a fast
  adaptive boundary attack. In: International Conference on Machine Learning.
  pp. 2196--2205. PMLR (2020)

\bibitem{croce2020reliable}
Croce, F., Hein, M.: Reliable evaluation of adversarial robustness with an
  ensemble of diverse parameter-free attacks. In: International conference on
  machine learning. pp. 2206--2216. PMLR (2020)

\bibitem{deng2009imagenet}
Deng, J., Dong, W., Socher, R., Li, L.J., Li, K., Fei-Fei, L.: Imagenet: A
  large-scale hierarchical image database. In: 2009 IEEE conference on computer
  vision and pattern recognition. pp. 248--255. Ieee (2009)

\bibitem{dieleman2016exploiting}
Dieleman, S., De~Fauw, J., Kavukcuoglu, K.: Exploiting cyclic symmetry in
  convolutional neural networks. In: International conference on machine
  learning. pp. 1889--1898. PMLR (2016)

\bibitem{dummit2004abstract}
Dummit, D.S., Foote, R.M.: Abstract algebra, vol.~3. Wiley Hoboken (2004)

\bibitem{madrylabrobustness}
Engstrom, L., Ilyas, A., Salman, H., Santurkar, S., Tsipras, D.: Robustness
  (python library) (2019), \url{https://github.com/MadryLab/robustness}

\bibitem{engstrom2019exploring}
Engstrom, L., Tran, B., Tsipras, D., Schmidt, L., Madry, A.: Exploring the
  landscape of spatial robustness. In: International conference on machine
  learning. pp. 1802--1811. PMLR (2019)

\bibitem{finzi2020generalizing}
Finzi, M., Stanton, S., Izmailov, P., Wilson, A.G.: Generalizing convolutional
  neural networks for equivariance to lie groups on arbitrary continuous data.
  In: International Conference on Machine Learning. pp. 3165--3176. PMLR (2020)

\bibitem{gens2014deep}
Gens, R., Domingos, P.M.: Deep symmetry networks. Advances in neural
  information processing systems  \textbf{27} (2014)

\bibitem{goodfellow2009measuring}
Goodfellow, I., Lee, H., Le, Q., Saxe, A., Ng, A.: Measuring invariances in
  deep networks. Advances in neural information processing systems  \textbf{22}
  (2009)

\bibitem{goodfellow6572explaining}
Goodfellow, I.J., Shlens, J., Szegedy, C.: Explaining and harnessing
  adversarial examples. arXiv preprint arXiv:1412.6572  (2014)

\bibitem{he2016deep}
He, K., Zhang, X., Ren, S., Sun, J.: Deep residual learning for image
  recognition. In: Proceedings of the IEEE conference on computer vision and
  pattern recognition. pp. 770--778 (2016)

\bibitem{higgins2022symmetry}
Higgins, I., Racani{\`e}re, S., Rezende, D.: Symmetry-based representations for
  artificial and biological general intelligence. Frontiers in Computational
  Neuroscience p.~28 (2022)

\bibitem{kayhan2020translation}
Kayhan, O.S., Gemert, J.C.v.: On translation invariance in cnns: Convolutional
  layers can exploit absolute spatial location. In: Proceedings of the IEEE/CVF
  Conference on Computer Vision and Pattern Recognition. pp. 14274--14285
  (2020)

\bibitem{kohler2020equivariant}
K{\"o}hler, J., Klein, L., No{\'e}, F.: Equivariant flows: exact likelihood
  generative learning for symmetric densities. In: International conference on
  machine learning. pp. 5361--5370. PMLR (2020)

\bibitem{krizhevsky2009learning}
Krizhevsky, A., Hinton, G.: Learning multiple layers of features from tiny
  images. Master's thesis, University of Toronto (2009)

\bibitem{krizhevsky2009cifar}
Krizhevsky, A., Nair, V., Hinton, G.: Cifar-10 and cifar-100 datasets. URl:
  https://www. cs. toronto. edu/kriz/cifar. html  \textbf{6} (2009)

\bibitem{krizhevsky2012imagenet}
Krizhevsky, A., Sutskever, I., Hinton, G.E.: Imagenet classification with deep
  convolutional neural networks. Advances in neural information processing
  systems  \textbf{25} (2012)

\bibitem{kurakin2016adversarial}
Kurakin, A., Goodfellow, I., Bengio, S.: Adversarial machine learning at scale.
  arXiv preprint arXiv:1611.01236  (2016)

\bibitem{lecun1995convolutional}
LeCun, Y., Bengio, Y., et~al.: Convolutional networks for images, speech, and
  time series. The handbook of brain theory and neural networks
  \textbf{3361}(10), ~1995 (1995)

\bibitem{lecun1989backpropagation}
LeCun, Y., Boser, B., Denker, J.S., Henderson, D., Howard, R.E., Hubbard, W.,
  Jackel, L.D.: Backpropagation applied to handwritten zip code recognition.
  Neural computation  \textbf{1}(4),  541--551 (1989)

\bibitem{lecun1998mnist}
LeCun, Y., Cortes, C., Burges, C.J.: The mnist database of handwritten digits,
  1998. URL http://yann. lecun. com/exdb/mnist  \textbf{10}, ~34 (1998)

\bibitem{lindqvist2022novel}
Lindqvist, B.: A novel method for function smoothness in neural networks. IEEE
  Access  \textbf{10},  75354--75364 (2022)

\bibitem{madry2017towards}
Madry, A., Makelov, A., Schmidt, L., Tsipras, D., Vladu, A.: Towards deep
  learning models resistant to adversarial attacks. arXiv preprint
  arXiv:1706.06083  (2017)

\bibitem{marcos2017rotation}
Marcos, D., Volpi, M., Komodakis, N., Tuia, D.: Rotation equivariant vector
  field networks. In: Proceedings of the IEEE International Conference on
  Computer Vision. pp. 5048--5057 (2017)

\bibitem{miller1973symmetry}
Miller, W.: Symmetry groups and their applications. Academic Press (1973)

\bibitem{moosavi2016deepfool}
Moosavi-Dezfooli, S.M., Fawzi, A., Frossard, P.: Deepfool: a simple and
  accurate method to fool deep neural networks. In: Proceedings of the IEEE
  conference on computer vision and pattern recognition. pp. 2574--2582 (2016)

\bibitem{art2018}
Nicolae, M.I., Sinn, M., Tran, M.N., Buesser, B., Rawat, A., Wistuba, M.,
  Zantedeschi, V., Baracaldo, N., Chen, B., Ludwig, H., Molloy, I., Edwards,
  B.: Adversarial robustness toolbox v1.0.1. CoRR  \textbf{1807.01069} (2018),
  \url{https://arxiv.org/pdf/1807.01069}

\bibitem{NEURIPS2019_9015}
Paszke, A., Gross, S., Massa, F., Lerer, A., Bradbury, J., Chanan, G., Killeen,
  T., Lin, Z., Gimelshein, N., Antiga, L., Desmaison, A., Kopf, A., Yang, E.,
  DeVito, Z., Raison, M., Tejani, A., Chilamkurthy, S., Steiner, B., Fang, L.,
  Bai, J., Chintala, S.: Pytorch: An imperative style, high-performance deep
  learning library. In: Wallach, H., Larochelle, H., Beygelzimer, A.,
  d\textquotesingle Alch\'{e}-Buc, F., Fox, E., Garnett, R. (eds.) Advances in
  Neural Information Processing Systems 32, pp. 8024--8035. Curran Associates,
  Inc. (2019),
  \url{http://papers.neurips.cc/paper/9015-pytorch-an-imperative-style-high-performance-deep-learning-library.pdf}

\bibitem{romero2020group}
Romero, D.W., Cordonnier, J.B.: Group equivariant stand-alone self-attention
  for vision. In: International Conference on Learning Representations (2020)

\bibitem{russakovsky2015imagenet}
Russakovsky, O., Deng, J., Su, H., Krause, J., Satheesh, S., Ma, S., Huang, Z.,
  Karpathy, A., Khosla, A., Bernstein, M., et~al.: Imagenet large scale visual
  recognition challenge. International journal of computer vision
  \textbf{115}(3),  211--252 (2015)

\bibitem{schmidt2012learning}
Schmidt, U., Roth, S.: Learning rotation-aware features: From invariant priors
  to equivariant descriptors. In: 2012 IEEE Conference on Computer Vision and
  Pattern Recognition. pp. 2050--2057. IEEE (2012)

\bibitem{sifre2013rotation}
Sifre, L., Mallat, S.: Rotation, scaling and deformation invariant scattering
  for texture discrimination. In: Proceedings of the IEEE conference on
  computer vision and pattern recognition. pp. 1233--1240 (2013)

\bibitem{sokolic2017generalization}
Sokolic, J., Giryes, R., Sapiro, G., Rodrigues, M.: Generalization error of
  invariant classifiers. In: Artificial Intelligence and Statistics. pp.
  1094--1103. PMLR (2017)

\bibitem{szegedy2013intriguing}
Szegedy, C., Zaremba, W., Sutskever, I., Bruna, J., Erhan, D., Goodfellow,
  I.J., Fergus, R.: Intriguing properties of neural networks. In: International
  Conference on Learning Representations (2013)

\bibitem{tramer2020adaptive}
Tramer, F., Carlini, N., Brendel, W., Madry, A.: On adaptive attacks to
  adversarial example defenses. Advances in Neural Information Processing
  Systems  \textbf{33},  1633--1645 (2020)

\bibitem{zhang2019making}
Zhang, R.: Making convolutional networks shift-invariant again. In:
  International conference on machine learning. pp. 7324--7334. PMLR (2019)

\bibitem{zhou2017oriented}
Zhou, Y., Ye, Q., Qiu, Q., Jiao, J.: Oriented response networks. In:
  Proceedings of the IEEE Conference on Computer Vision and Pattern
  Recognition. pp. 519--528 (2017)

\end{thebibliography}
%

\end{document}